# Improved Tissue Sodium Concentration Quantification in Breast Cancer by Reducing Partial Volume Effects – a Preliminary Study

Olgica Zaric[1,2], Carmen Leser[3], Vladimir Juras[4], Alex Farr[3], Pavol Szomolanyi[4], Malina Gologan[4], Stanislas Rapacchi[5,6,7], Laura Villazan Garcia[1], Haider Ali[1], Christian Singer[3], Siegfried Trattnig[2,4], Christian Licht[8,9,10*], Ramona Woitek[1*]

[1]Research Centre for Medical Imaging and Image Analysis (MIAAI), Danube Private University, Krems, Austria

[2]Institute for Musculoskeletal Imaging, Karl Landsteiner Society, St. Pölten, Austria

[3]Department for Gynecology and Obstetrics, Medical University of Vienna, Vienna, Austria

[4]High-field MR Centre, Medical University of Vienna, Vienna, Austria

[5]Department of Radiology, Lausanne University Hospital and University of Lausanne, Lausanne, Switzerland

[6]Aix-Marseille University, CNRS, CRMBM, Marseille, France

[7]APHM, Hôpital Universitaire Timone, CEMEREM, Marseille, France

[8]Radiological Sciences Laboratory, School of Medicine, Stanford University, Stanford, California, USA

[9]Computer Assisted Clinical Medicine, Medical Faculty Mannheim, Heidelberg University, Mannheim, Germany

[10]Mannheim Institute for Intelligent Systems in Medicine, Medical Faculty Mannheim, Germany

*Authors equally contributed to this manuscript

**Corresponding author: Olgica Zaric

E-mail: olgica.zaric@dp-uni.ac.at

Address: Research Centre for Medical Imaging and Image Analysis (MIAAI), Danube Private University (DPU), Viktor Kaplan-Straße 2, 2700 Wiener Neustadt, Austria

**Abstract:**

**Introduction:** In sodium ($^{23}$Na) magnetic resonance imaging (MRI), partial volume effects (PVE) are one of the most common causes of errors in the in vivo quantification of tissue sodium concentration (TSC). Advanced image reconstruction algorithms, such as compressed sensing (CS), have the potential to reduce PVE. Therefore, we investigated the feasibility of using CS-based methods to improve image quality and TSC quantification accuracy in patients with breast cancer.

**Subjects and methods:** In this study, three healthy participants and 12 female participants with breast cancer were examined on a 7T MRI scanner. $^{23}$Na-MRI images were reconstructed using weighted total variation (wTV), directional total variation (dTV), anatomically guided total variation (AG-TV) and adaptive combine (ADC) methods. The consistency of tumor volume delineations based on sodium data was assessed using the Dice score, and TSC quantification was performed for various image reconstruction methods. Pearson's correlation coefficients were calculated to assess the relationships between wTV, dTV, AG-TV, and ADC values.

**Results:** All methods provided breast MRI images with well-preserved sodium signal and tissue structures. The mean Dice scores for wTV, dTV, and AG-TV were 65±11%, 72±9%, and 75±6%, respectively. Average TSC values in breast tumors were 61.0±3.0, 72.0±3.0, 73.0±4.0, and 88.0±4.0 mmol/L for wTV, dTV, AG-TV, and ADC, respectively. A strong negative correlation was observed between wTV and dTV ($r = -0.78$, 95% CI [−0.94, −0.31], $p = 0.0076$) and a strong positive correlation between dTV and AG-TV ($r = 0.71$, 95% CI [0.16, 0.92], $p = 0.0207$) was found.

**Conclusion:** The results of this study showed that differences in tumor appearance and TSC estimations may depend on the type of image reconstruction and the parameters used. This is most likely due to differences in their ability to reduce PVE.

**Abbreviations:**

ADC, adaptive combine

AG-TV, anatomically guided total variation

$B_0$, static magnetic field

$B_1$, combined transmit and receive radiofrequency field

$B_1^+$, transmit radiofrequency field

CS, compressed sensing

dTV, directional total variation

PVE, partial volume effect

SNR, signal to noise ratio

TSC, tissue sodium concentration

wTV, weighted total variation

## 1 Introduction

Sodium magnetic resonance imaging ($^{23}$Na-MRI) can provide non-invasive biochemical information on cell viability, structural integrity, and energy metabolism. This enables the assessment of pathophysiological changes in human tissue [1, 2]. One of the most promising imaging markers for identifying breast tumors and evaluating treatment responses is the quantitative parameter derived from $^{23}$Na-MRI data: tissue sodium concentration (TSC) [3-7].

However, due to the low nominal resolution of sodium imaging, partial volume effects (PVE) can substantially impact the accuracy of TSC quantification. The most common nominal isotropic resolution for breast $^{23}$Na-MRI is 3 mm [8, 9], which implies a relatively large voxel size. Consequently, PVE become more pronounced, resulting in increased mixing of signals between different tissue types and reduced quantitative accuracy. To mitigate PVE, partial volume correction (PVC) techniques, such as the algorithm proposed by Niesporek et al., are employed. This method accounts for point-spread function (PSF) blurring and voxel resolution limitations to adjust the measured sodium signal, thereby reducing PVE-related bias in TSC measurements [10]. Additionally, Kim et al. proposed a voxel-wise PVC method for $^{23}$Na-MRI, which incorporated anatomical priors to refine TSC measurements. This method demonstrated improved accuracy in TSC estimation by accounting for tissue fraction effects and PSF blurring [11].

Compressed sensing (CS) reconstruction has the potential to mitigate partial volume effects by enhancing spatial fidelity in sodium MRI through the exploitation of image sparsity in the gradient or transform domain [12, 13]. Since tissue boundaries are naturally sparse, enforcing sparsity during reconstruction can effectively restore sharp edges and minimize blurring caused by the finite point spread function (PSF) associated with limited k-space sampling [13]. This approach permits smaller effective voxel sizes for the same signal-to-noise-ratio (SNR) and acquisition time, improving the depiction of fine tissue structures without prolonging acquisition. Furthermore, CS regularisation suppresses noise while preserving boundary information. This is particularly important in sodium MRI, where the low intrinsic SNR and coarse resolution can obscure the interfaces between distinct tissue types. Building upon the principles of sparsity and anatomical guidance, several groups have extended CS and model-based reconstruction (MBR) frameworks using prior structural information from proton MRI. Lachner et al. demonstrated that incorporating anatomical priors from proton MR contrasts through anatomically weighted first- and second-order total variation (TV) enhances the apparent resolution of known structures [14, 15]. Zhao et al. further proposed a motion-compensated generalized series reconstruction combined with a sparse anatomical prior, where segmented, high-resolution proton images were employed for denoising and resolution enhancement. Their results showed that the anatomically constrained reconstruction substantially improved both SNR and lesion fidelity [16]. Ehrhardt et al. advanced this concept by introducing a segmentation-free directional total variation (dTV) framework that integrates anatomical structure information directly into the reconstruction process. This method achieves simultaneous denoising and resolution enhancement of $^{23}$Na images and enables local signal-decay estimation and compensation through a joint dual-echo reconstruction approach [17]. More recently, Licht et al. extended the dTV framework to the multi-quantum coherence (MQC) domain of sodium MRI, employing $^{1}$H-derived tissue-specific priors to guide reconstruction and separate overlapping tissue signals. This strategy further mitigates partial volume effects by combining compressed sensing, model-based fitting, and anatomical constraints, leading to sharper structural boundaries, improved quantitative accuracy, and more reliable visualization of tissue interfaces [18].

In this study, we aimed to evaluate the feasibility of the reconstruction methods proposed by Ehrhardt et al. (method #1) and Licht et al. (method #2) for improving tissue sodium concentration quantification in breast MRI. We further compared these approaches with TSC values derived from $^{23}$Na-MR images reconstructed using the adaptive combined (ADC) method [19]. Our goal was to determine whether these advanced reconstruction methods can improve the delineation of fine tissue structures and reduce PVE, thereby enabling more accurate quantification of TSC in breast tumors [19].

## 2 Materials and Methods

### 2.1 Calibration probes and calibration phantom

Prior to the start of the study, two vials containing a saline solution (NaCl) with sodium concentrations of 77 and 154 mmol/L, mixed with 4% agarose gel (Agarose, Sigma-Aldrich, USA), were prepared, and attached to the inner parts of both breast coil elements. The mixture was treated with 2.9 g/L of copper sulphate to match the T1 relaxation times of breast tissue. A phantom was created by assigning TSC values of 40 and 20 mmol/L to glandular and adipose tissue respectively, and 77 and 154 mmol/L to the two vials. This phantom served as a dual reference, enabling signal intensities to be read from the glandular and adipose tissue and the accuracy of the vial concentrations to be evaluated for the two tissue types, and vice versa.

### 2.2 Study participants

This study enrolled three healthy female subjects and 12 women with breast cancer. All MRI scans were performed between January 2023 and March 2024 at the Medical University of Vienna and were approved by the local research ethics committee (approval number: 1131/2015; cooperation agreement number: 030-2023). Written informed consent was obtained from all participants. All methods were performed in accordance with the Declaration of Helsinki.

### 2.3 MR imaging

MR imaging was performed using a whole-body 7.0 T MRI scanner with a gradient amplitude of 70 mT/m and a slew rate of 200 T/m/s (Magnetom Dot Plus, Siemens Healthineers, Germany).

All participants were scanned in the prone position. A Tx/Rx dual-tuned $^{23}$Na and $^{1}$H bilateral breast coil (14 $^{23}$Na channels and two $^{1}$H channels; QED, Cleveland, USA) was used.

$^{23}$Na MRI data sets were acquired using a density-adapted three-dimensional radial projection reconstruction pulse sequence (3D DA-PR) [20]. For a nominal spatial resolution of 3 mm, the sequence parameters were as follows: repetition time (TR) and echo time (TE) were 100 ms and 0.55 ms, respectively; pulse duration was 1 ms; the nominal flip angle of 90° was calibrated for maximum signal intensity; the readout time was 10.02 ms; the bandwidth was 100 Hz/px; and the acquisition time was 16 minutes. The number of projections was 8000, with 384 samples per projection. Sodium imaging was followed by noise-only scans (no radiofrequency power). Morphological and proton ($^{1}$H) three-dimensional double-echo steady-state (3D DESS) [21] images were acquired with the following parameters: field of view: 320 × 320 mm²; nominal resolution: 1 mm³; TR: 9.3 ms; TE: 2.6 ms; and acquisition time: 2:30 min. The total scanning time for this study was around 20 minutes.

**2.4 Breast tissue and tumor segmentation**

Different tissues were segmented using high-resolution $^{1}$H DESS images. In healthy subjects, the glandular and adipose tissues, as well as the vials containing 77 mmol/L and 154 mmol/L solutions, were labelled as four different compartments. In patients, the entire breast, tumor and calibration phantoms were manually segmented by two non-clinical readers using ITK-SNAP software (version 4.0.0), under the supervision of a breast radiologist who had received subspecialty training. As high-resolution morphological $^{1}$H MR imaging was performed at 7T without contrast medium, 3T data were used for precise lesion location and margin verification. The same segmentation mask was applied to $^{23}$Na-MRI images reconstructed using different methods.

**2.5 Image reconstruction**

$^{23}$Na-MRI images were reconstructed offline using custom-written scripts in MATLAB software (version R2022a; MathWorks, Natick, USA).

**2.5.1 Weighted Total Variation (wTV) and Directional Total Variation (dTV)**

Ehrhardt et al. proposed two structurally guided priors—weighted total variation (wTV) and directional total variation (dTV)—to improve multi-contrast MRI reconstruction by leveraging anatomical edge information from one contrast to inform another. Both priors are convex, enabling the use of convex optimization algorithms, and the reconstruction is performed using the alternating direction method of multipliers (ADMM) with a fast gradient projection method to efficiently compute the proximal operator [17]. By maintaining clear anatomical boundaries, wTV and dTV can minimize signal interference between different types of tissue and enhance the precision of TSC measurements.

### 2.5.2 Anatomically Guided Total Variation (AG-TV)

The AG-TV method reconstructs high-quality $^{23}$Na MRI images by incorporating structural information from high-resolution $^{1}$H images. It aims to recover missing high-frequency components and reduce PVE through anatomically informed regularization. Edge information from the $^{1}$H images is extracted using first-order derivatives along all spatial directions, which are inverted to create maps that locally enhance edges and guide the total variation (TV) denoising. The contribution of the prior can be adjusted using a threshold parameter. An additional regularization term suppresses background regions outside the tissue of interest, further reducing PVE. The reconstruction problem is solved using the Split-Bregman algorithm, with empirically determined weighting factors to balance the influence of edge enhancement and background suppression [18, 22].

## 2.6 Image quality assessments

We used the Dice score to compare tumour volumes segmented on sodium images, and we analyzed the signal intensity profiles (ImageJ, LOCI, University of Wisconsin, USA) for each patient. This is because the distribution of signal intensity is determined by the point spread function, which affects the level of detail and sharpness in the image. Image quality was assessed using the structural similarity index (SSIM) and root mean square error (RMSE) [23] and the focus measure (FM) [24], which is a variation of the Laplacian. This determined whether the reconstructed images were less blurred than those reconstructed using the ADC approach.

**2.7 TSC quantification**

The signal intensities of the calibration solutions obtained from the $^{23}$Na-MR images reconstructed using methods 1, 2 and reference were fitted to a two-point calibration curve [25]. The signals for each segmentation region were automatically provided by the software. For the glandular tissue, we included healthy subjects and the contralateral breast of patients. The sodium signals from the vials were used to derive a calibration curve and determine TSC values. This approach assumed that the water concentration was the same in the standards (100%) as in the tissue (~75%), and therefore a correction factor of 0.75 was applied to account for the lower water content in tissue, to avoid underestimating TSC in breast tumours [8].

**2.8 Statistical tests**

The data were analyzed using the IBM SPSS Statistics software package (version 30.0, Armonk, NY, USA). Metric data are presented as means (mean ± SD), and the normal distribution of the measured values was tested using the Shapiro–Wilk test. For categorical data, absolute frequencies and percentages were presented. The relationship between image metrics was analyzed using the paired samples t-test. A Pearson correlation test was performed to analyse the consistency of TSC results obtained using different methods in glandular tissue and tumors. A p-value smaller than 0.05 was considered significant.

**3 Results**

The mean age of the three healthy subjects was 40.0 ± 8.0 years, compared to 55.0 ± 14.0 years for the cancer patients. All data showed a normal distribution. Eleven participants were diagnosed with invasive ductal carcinoma (IDC) and one with invasive lobular carcinoma (ILC). The mean one-dimensional tumor size was 16±8 mm. For patients with multiple tumors, the largest tumour was included in this analysis. Further details on the demographic data of the study participants with breast tumours can be found in Table 1.

### 3.1 Image quality evaluation

The images showed that methods 1 and 2 produced images with well-retained sodium signal intensities and tissue structures (Figure 1 (a-l), Figure 2 (a-l), Figure 3 (a-l)). However, lesion conspicuity was not the same for images reconstructed using wTV, dTV and AG-TV. For method#1, the tumor margins were better visualized with increasing number of projections; differences in lesion volume and shape were observed compared to proton images. When comparing tumor segmentation of ground truth images with method#1, we found a Dice score of 65±11% for wTV and 72±6% and for dTV. The same comparison gave a Dice score of 75±8% for AG-TV. When we analyzed the signal intensity profile through the tumour, we found differences in their values and shapes, as shown in the example of one subject (Figure 4 (a-j)).

The mean±SD of quantitative parameters (SSIM, RMSE, and FM) for two methods were calculated for each patient. As expected, method #1 showed that improving the number of radial projections from eight to 64 enhanced image quality. For 64 radial projections method#1 (wTV)), SSIM was 0.39±0.02, RMSE was 45.60±5.0, and FM was 46.6±9.0. For the same parameters, the dTV method provided the following values: RMSE was 45.50±4.0; SSIM was 0.39±0.02; and FM was 62.2±4.0. When SSIM between wTV and dTV were compared, we found no significant difference for each number of projections: for 8 projections p=0.904; for 16 projections, p=0.380; and p=0.342 and p=0.228 for 32 and 64 projections, respectively (Figure 5). There were similar findings for RMSE: p=0.530, p=0.065, p=0.865, and p=0.964, for 8, 16, 32, and 64 projections, respectively. However, when we compared the average focus measure between wTV vs dTV, we found significant differences for 8, 16, 32, and 64 projections ($p<0.001$). For AG-TV, changing number of iterations did not influence image quality metrics. SSIM, RMSE, and FM were 0.40±0.04, 41.6±5.0, and 30.0±4.0, respectively. For the reference method, the FM was 26.3±5.0, which was 43.5% lower than wTV (64 projections), 57.9% lower than dTV (64 projections) and 12.3% lower than AG-TV.

When dTV vs AG-TV was compared in terms of SSIM, RMSE, and FM, the differences were as follows: $p=0.472$, $p=0.004$, and $p<0.001$, respectively (Figure 5).

### 3.2 TSC quantitative evaluations

#### 3.2.1 Phantom measurements

The TSC values for the phantoms with known concentrations of 77.0 and 154.0 mmol/L were calculated using all three methods and are shown below. For wTV (dTV), the TSC values were 88.0 ± 12.0 (80.0 ± 6.0) and 162.0 ± 11.0 (159.0 ± 9.0) mmol/L for 77.0 and 154.0 mmol/L, respectively. For AG-TV, the values were 88.0 ± 10.0 and 168.0 ± 7.0, respectively. For the reference method, the values were 89.0 ± 5.0 mmol/L and 165.0 ± 10.0 mmol/L, respectively.

#### 3.2.2 *In vivo* measurements

The TSC in healthy glandular tissue was 40.0±2.0, 42.0±2.0 mmol/L, 50.0±2.0 mmol/L, 57.0±2.0 mmol/L, whereas in tumors it was 61.0±3.0 mmol/L, 72.0±3.0 mmol/L, 73.0±4.0 mmol/L and 88.0±4.0 mmol/L, as evaluated by wTV, dTV, AG-TV and the reference method, respectively (Figure 6). In glandular issue, the correlation between AG-TV and ADC was the strongest ($r=0.74$, 95% CI [0.36, 0.91], $p=0.002$), wTV and dTV also showed moderate positive correlations with ADC ($r=0.62$, 95% CI [0.12, 0.86], $p=0.01$ and $r=0.59$, 95% CI [0.08, 0.85], $p=0.02$ respectively). A moderate positive correlation was observed between dTV and AG-TV values ($r=0.50$, 95% CI [−0.02, 0.80], $p=0.06$). In tumors, a strong negative correlation was observed between wTV and dTV ($r = -0.78$, 95% CI [−0.94, −0.31], $p = 0.0076$, and a strong positive correlation was found between dTV and AG-TV ($r = 0.71$, 95% CI [0.16, 0.92], $p = 0.0207$) (Figure 7).

## 4 Discussion

In this study, we implemented and evaluated the feasibility of the image reconstruction methods proposed by Ehrhardt et al. and Licht et al. in breast tissue, focusing on both image quality and TSC quantification accuracy [15, 16]. These methods leverage structural information from high-resolution anatomical MRI to guide $^{23}$Na-MRI reconstruction, aiming to preserve tissue boundaries and reduce PVE. For comparison, $^{23}$Na-MR images reconstructed using an ADC algorithm were used to benchmark improvements in anatomical detail and quantitative accuracy. Both method #1 (wTV, dTV) and method #2 (AG-TV) produced $^{23}$Na-MRI images with well-preserved sodium signal and tissue structures. Lesion conspicuity varied across methods: tumor margins improved with increasing radial projections for dTV, and Dice scores were highest for AG-TV. Quantitative metrics (SSIM, RMSE, FM) generally improved with more projections for wTV and dTV, whereas AG-TV metrics remained stable across iterations. Focus measure analysis indicated better edge definition for dTV compared to wTV.

Previous studies have highlighted that image quality and the availability of high-quality multiparametric MRI data are crucial for accurate tumor delineation [26, 27]. In our study, the applied reconstruction methods produced highly detailed and visually consistent $^{23}$Na-MRI images. However, we observed notable discrepancies in tumor morphology, including differences in shape and volume, when comparing wTV-, dTV-, and AG-TV–reconstructed images with the reference. These variations likely reflect the influence of structural priors and regularization constraints, which can enhance edge definition and suppress noise but may also alter boundary appearance and affect volumetric accuracy, particularly in regions with low intrinsic sodium signal and low SNR. Moderate Dice scores indicated differing levels of robustness among the methods in faithfully representing tumor margins. Reconstructions also varied in the depiction of small morphological details, tumor borders, and skin visibility, demonstrating that tissue visualization fidelity depends strongly on the chosen reconstruction method. Analysis of signal intensity profiles, used to estimate PSF, further revealed considerable variation across methods. While SSIM did not show substantial differences, the greatest discrepancies in RMSE and FM were observed between dTV and AG-TV.

Although CS improves image quality and accelerates acquisition, its regularization (i.e, total variation, TV, or wavelet regularization, wTV) inherently imposes spatial smoothing, which can bias quantitative measurements [12]. Sparsity-enforcing priors (e.g., TV and wTV) suppress noise by penalizing intensity variations but may also attenuate signal in low-intensity or heterogeneous regions, potentially losing fine spatial details and physiologically relevant fluctuations. In $^{23}$Na-MRI, where intrinsic SNR is low and TSC differences are subtle, this smoothing can systematically underestimate TSC in high-signal regions and overestimate it in low-signal regions, reducing apparent tissue contrast. Blunck et al. demonstrated that increasing regularization progressively reduces spatial noise but biases TSC estimates, particularly in brain regions with fine structural variation [28]. Similarly, Utzschneider et al. and Behl et al. reported that while CS-based reconstructions improve apparent SNR, excessive regularization can smooth small structures and distort quantitative accuracy [29, 30].

Our study revealed differences in TSC through image quality metrics but incorporating proton anatomical information that is not present in the sodium dataset can introduce artificial structures. If these regions contain sodium-rich tissue, such as glandular tissue, TSC may be overestimated. Therefore, only reliable prior information—well-defined surrounding glandular and tumor tissue—should be used. However, in a large, homogeneous, dense glandular region with minimal PVE, dTV reconstructions showed good agreement with AG-TV (p=0.06), and the values obtained were in good agreement with literature data, suggesting that neither dTV nor AG-TV significantly underestimated the signal [31]. The observed significant difference between dTV and AG-TV in tumorous regions (p=0.0207) indicates that tumour-specific factors, including size, spatial position, and heterogeneity, likely contribute to variations in the TSC estimates. This finding highlights the sensitivity of AG-TV to local gradient and texture complexity, which may better reflect the structural diversity present within tumour tissue compared with the more directionally constrained dTV metric.

One limitation of this study is that $B_0$ and $B_1^+$ inhomogeneity corrections were not applied. However, it is expected that these effects will be similar across all reconstruction methods and will not affect relative comparisons. In this study, two patients with large tumours (i.e., greater than 2.0 cm in diameter) were included, while the remaining patients had small to moderate-sized lesions. Future studies should incorporate larger, more diverse cohorts to assess the influence of reconstruction methods on TSC quantification across tumours of varying sizes.

## 5 Conclusion

Advanced image reconstruction methods play a crucial role in reducing partial volume effects in $^{23}$Na-MRI by enhancing image quality, increasing SNR, and refining tissue boundary definition. Structural-prior–guided techniques, such as directional total variation and anatomy-guided total variation, effectively mitigate voxel mixing and improve tissue differentiation for more accurate TSC quantification. While the reconstruction approaches preserve overall image quality, observed variations in tumor morphology and TSC measurements emphasize the influence of regularization strength and prior information on quantitative accuracy. Tailoring the reconstruction strategy to the specific imaging modality and clinical application is therefore essential for minimizing PVE and optimizing quantitative $^{23}$Na-MRI performance.

**Conflict of interest statement:**

The authors declare that they have no known competing financial interests or personal relationships that could have appeared to influence the work reported in this paper.

**Table 1**: Demographic features of participants of the study

| Subject No. | Age (years) | BIRADS | Menopause status | Tumor size (mm) | Verification | Histologic type | Grade |
|---|---|---|---|---|---|---|---|
| 1 | 43 | / | Pre | / | / | / | / |
| 2 | 38 | / | Pre | / | / | / | / |
| 3 | 39 | / | Pre | / | / | / | / |
| 4 | 40 | 5 | Pre | 12 | Biopsy | IDC | 2 |
| 5 | 59 | 5 | Post | 35 | Biopsy | IDC | 3 |
| 6 | 58 | 4 | Post | 9 | Biopsy | IDC | 1 |
| 7 | 29 | 5 | Pre | 19 | Biopsy | IDC | 2 |
| 8 | 59 | 5 | Post | 15 | Biopsy | IDC | 2 |
| 9 | 71 | 2 | Post | 9 | Biopsy | IDC | 2 |
| 10 | 46 | 4 | Pre | 18 | Biopsy | IDC | 2 |
| 11 | 38 | 5 | Pre | 18 | Biopsy | IDC | 3 |
| 12 | 65 | 5 | Post | 26 | Biopsy | IDC | 3 |
| 13 | 72 | 4 | Post | 5 | Biopsy | IDC | 2 |
| 14 | 46 | 4 | Post | 20 | Biopsy | IDC | 3 |
| 15 | 71 | 4 | Post | 11 | Biopsy | ILC | 2 |

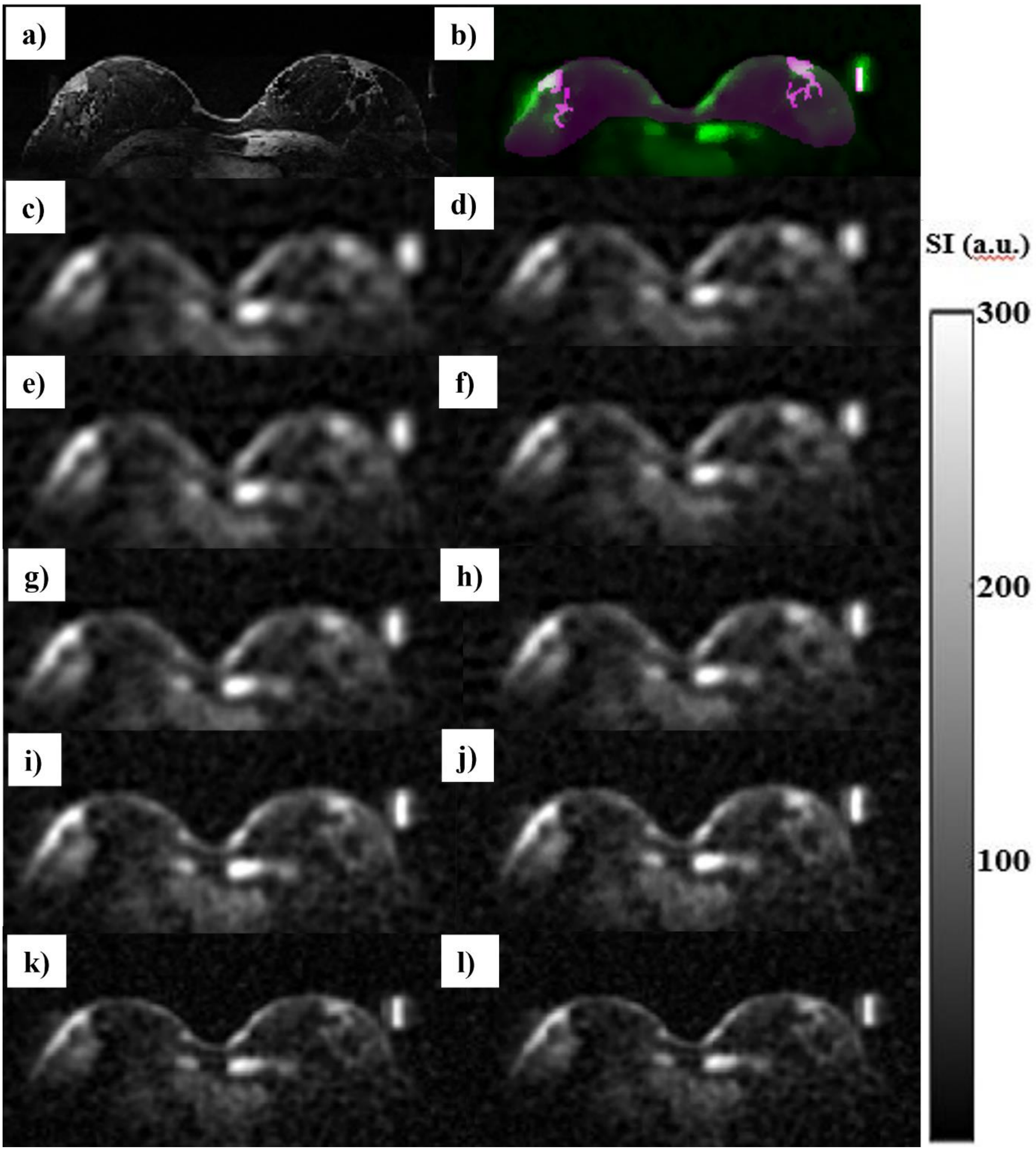

**Figure 1:** The figure demonstrates MR images of a 44-year-old healthy female subject (#1), a) proton DESS image in axial orientation, b) segmentation mask of breast, a sodium ($^{23}$Na) MR image reconstructed using wTV method for c) 8 projections, g) 16, e) 32, and i) 64 radial projections and dTV for d) 8 projections, f) 16, k) 32, and j) 64 radial projections. k) show the image reconstructed using AG-TV and a l) ground truth sodium image (ADC reconstruction). Gray-scale bar shows image signal intensity (SI) in arbitrary units (a.u.)

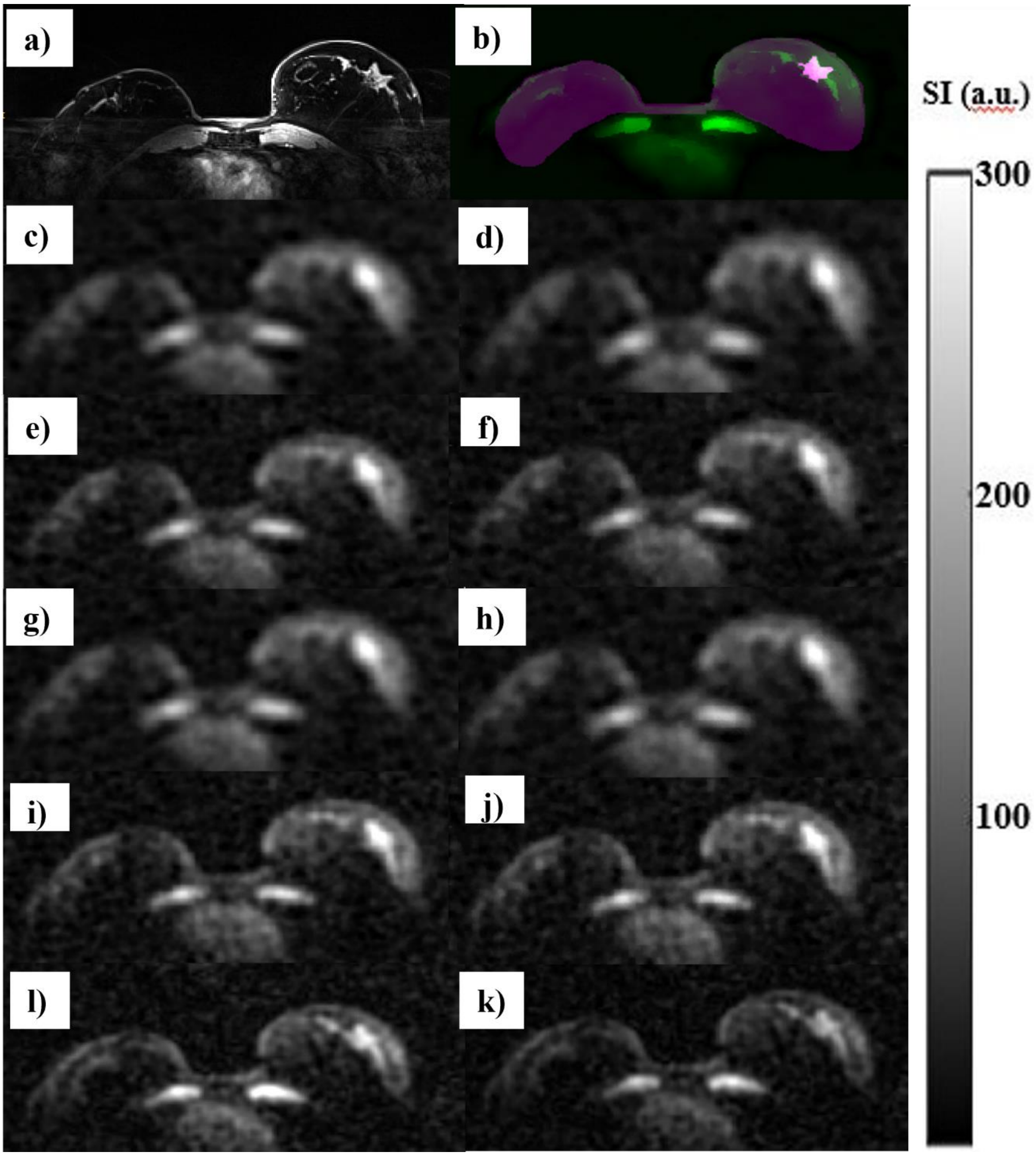

**Figure 2:** A 46-year-old female subject (#10) with invasive ductal carcinoma (IDC) of grade 2 in her left breast. The figure demonstrates a sodium ($^{23}$Na) MR image reconstructed using wTV for c) 8 projections, g) 16, e) 32, and i) 64 radial projections and dTV for d) 8 projections, f) 16, k) 32, and j) 64 radial projections. k) show the image reconstructed using the AG-TV and a l) ground truth sodium image (ADC). Gray-scale bar shows image signal intensity (SI) in arbitrary units (a.u.)

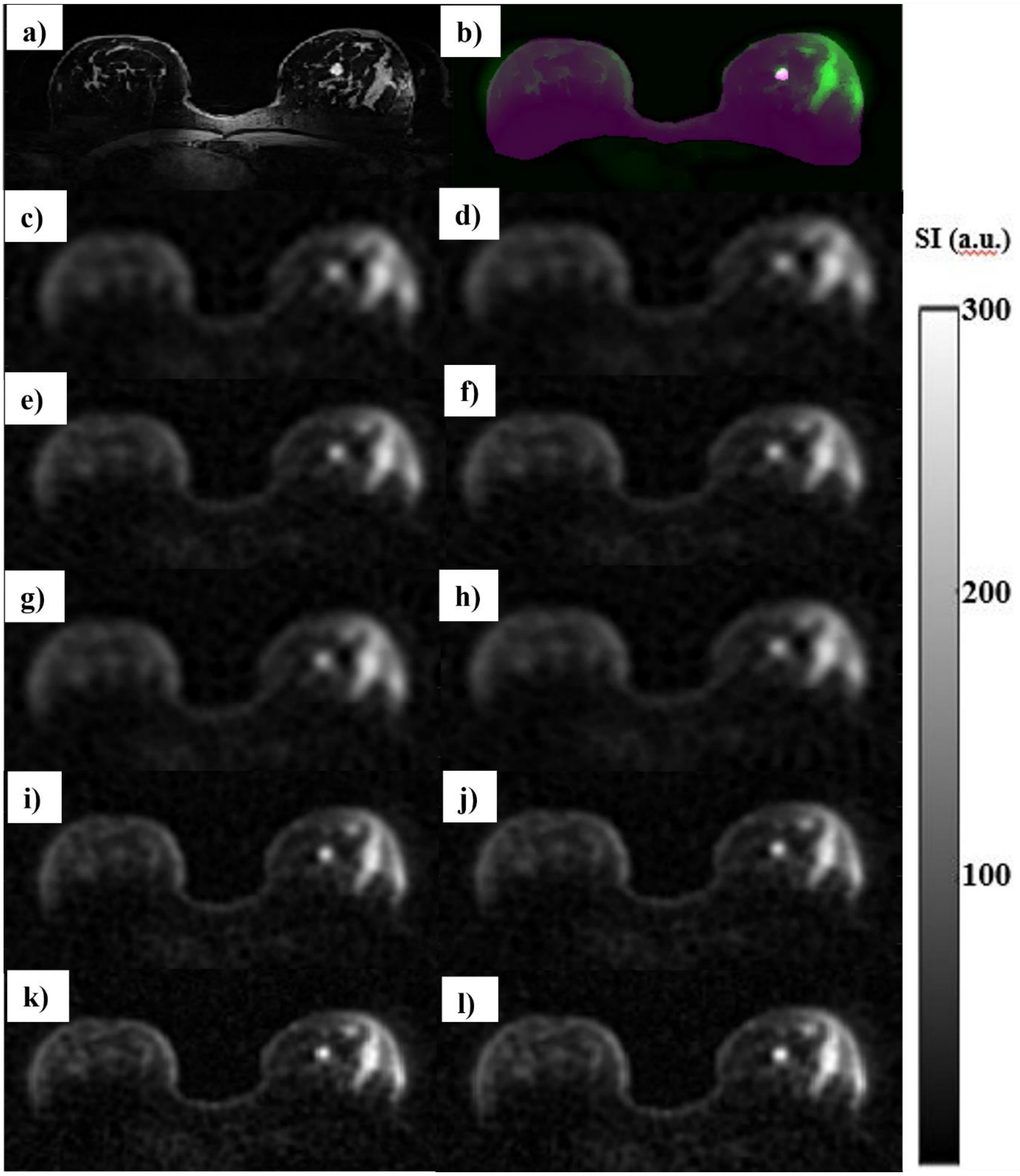

**Figure 3**: A 46-year-old female subject (#14) with multicentric invasive ductal carcinoma (IDC) of grade 3 in her left breast. The figure demonstrates a sodium ($^{23}$Na) MR image reconstructed using wTV for c) 8 projections, g) 16, e) 32, and i) 64 radial projections and dTV for d) 8 projections, f) 16, k) 32, and j) 64 radial projections. k) show the image reconstructed using the AG-TV and a l) ground truth sodium image (ADC). Gray-scale bar shows image signal intensity (SI) in arbitrary units (a.u.)

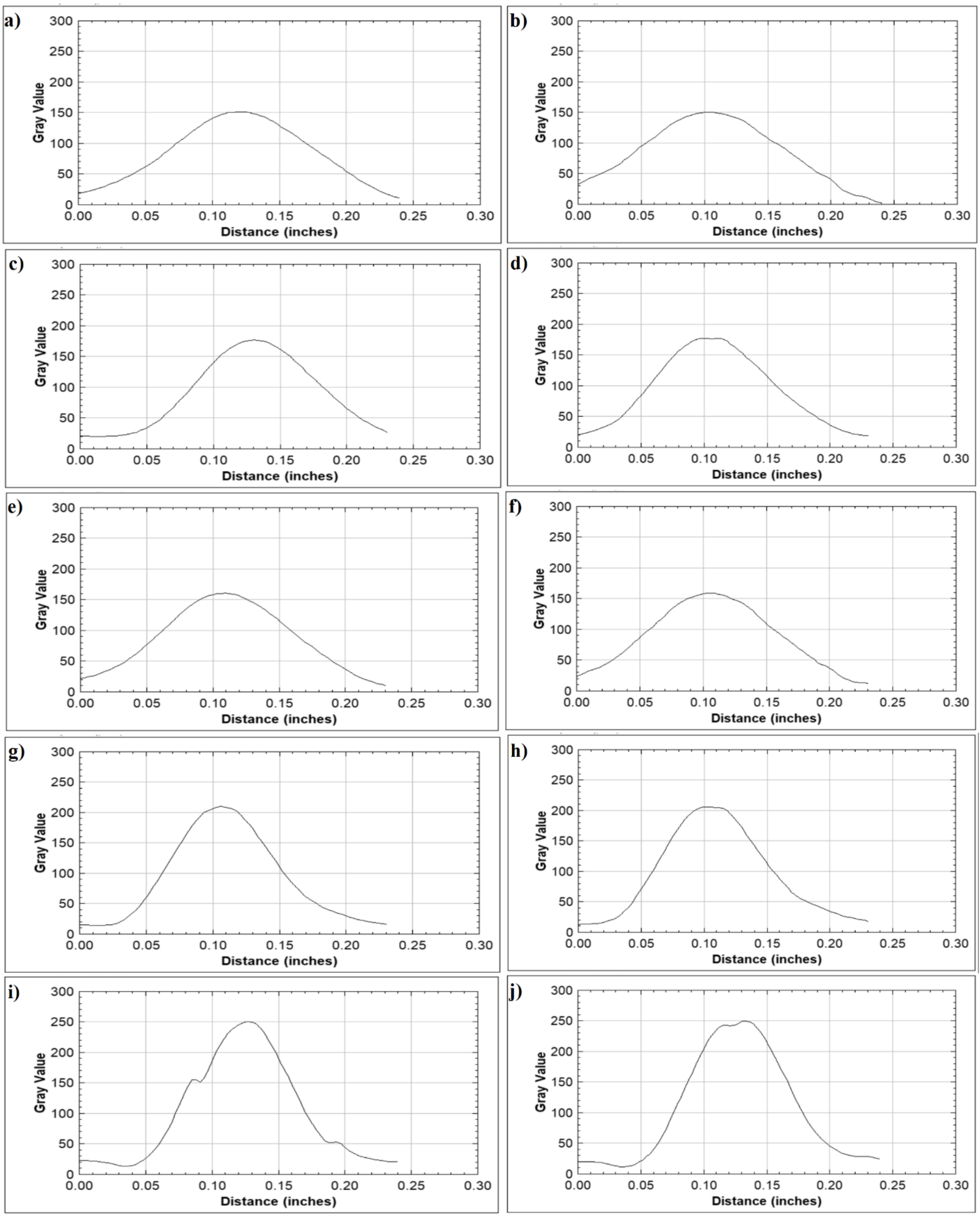

**Figure 4:** Graphical representation of the sodium signal intensity (SI) profiles through the tumor of the subject#14. The SI profiles were generated on $^{23}$Na-MR images reconstructed using wTV for a) 8 projections, c) 16, e) 32, and g) 64 radial projections and dTV for b) 8 projections, d) 16, f) 32, and h) 64 radial projections. i) show the image reconstructed using the AG-TV and a j) ADC reconstruction.

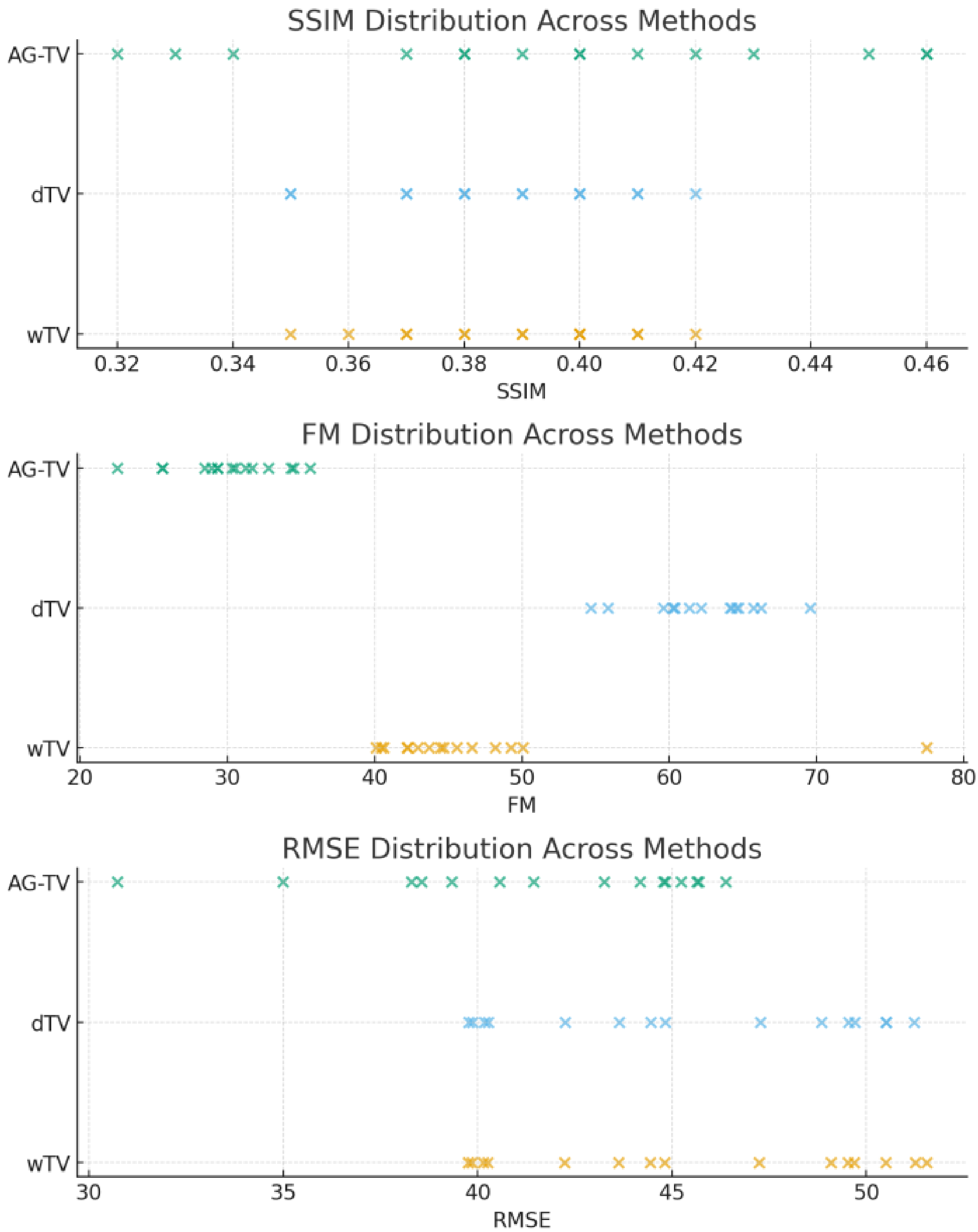

**Figure 5:** Distribution of image quality metrics across reconstruction methods. Horizontal dot plots show the distribution of SSIM, FM, and RMSE values for 64 projections for wTV (yellow), dTV (blue), and AG-TV (green) reconstruction methods. Each dot represents an individual subject.

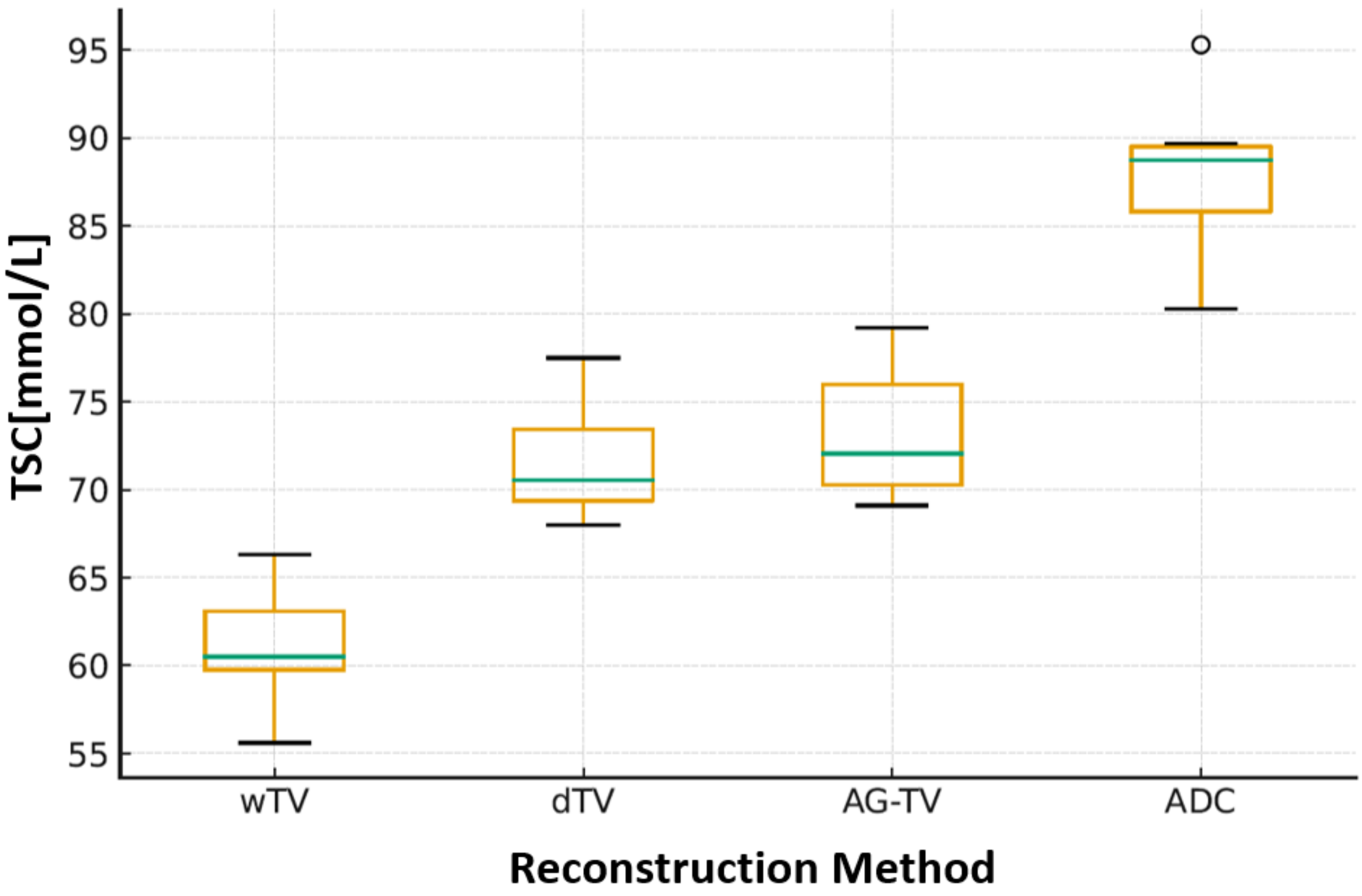

**Figure 6:** The boxplot diagram shows the range and median of the tissue sodium concentrations (TSC) found in breast cancers for the different reconstruction methods.

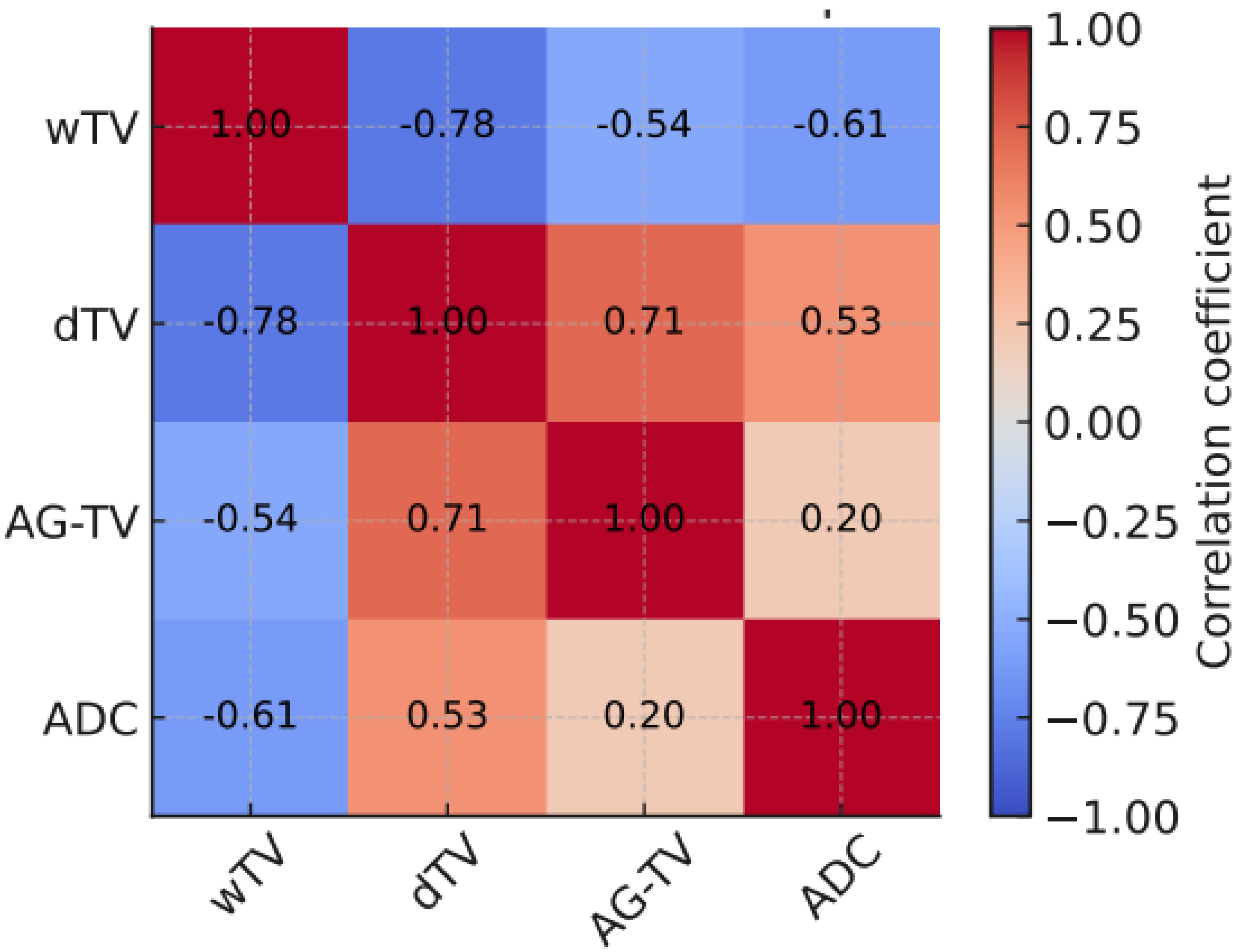

**Figure 7:** Pearson correlation heatmap of tissue sodium concentrations (TSC) from breast tumors across different reconstruction methods. Stronger correlations indicate greater agreement between methods.